# A new soft computing method for integration of experts' knowledge in reinforcement learning problems

Mohsen Annabestani[1], Ali Abedi[2], Mohammad Reza Nematollahi[1], and Mohammad Bagher Naghibi Sistani[3]

[1]Faculty of Electrical Engineering, Sharif University of Technology, Tehran, Iran
[2]School of Electrical Engineering, Iran University of Science and Technology, Tehran, Iran
[3]Department of Electrical Engineering, Ferdowsi University of Mashhad, Mashhad, Iran

Corresponding author: Mohsen Annabestani (e-mail: annabestany@ gmail.com).

**ABSTRACT** This paper proposes a novel fuzzy action selection method to leverage human knowledge in reinforcement learning problems. Based on the estimates of the most current action-state values, the proposed fuzzy nonlinear mapping assigns each member of the action set to its probability of being chosen in the next step. A user tunable parameter is introduced to control the action selection policy, which determines the agent's greedy behavior throughout the learning process. This parameter resembles the role of the temperature parameter in the softmax action selection policy, but its tuning process can be more knowledge-oriented since this parameter reflects the human knowledge into the learning agent by making modifications in the fuzzy rule base. Simulation results indicate that including fuzzy logic within the reinforcement learning in the proposed manner improves the learning algorithm's convergence rate, and provides superior performance.

## I. INTRODUCTION

Reinforcement learning (RL) has gained huge attention in the past decades. RL can be thought of as a kind of adaptation capability that enables the agent's optimal interaction with a complex and dynamic environment to solve sophisticated sequential decision-making problems [1]. RL is applied to various fields such as robotics, optimal and automatic control as in [2, 3]. RL applications in the semi-Markovian decision-making problems, stochastic games (competitive Markovian problems), hierarchical Markovian problems, and non-Markovian problems have also been reported in the literature [4-6].

RL algorithms have a close formulation to dynamic programming (DP) and frequently has been proposed as the solution of large-scale Markovian DP problems, in which the classical DP algorithms failed due to either curse of dimensionality or the curse of modeling [7]. A variety of RL formulation and algorithms have been developed based on different assumptions regarding the underlying problems or the techniques they have used to solve the problem. For example, [8, 9] have addressed deep learning-based approaches for solving RL problems named deep reinforcement learning (deep RL), in single- or multi-agent learning frameworks.. Authors in [10] surveyed the integration of natural language understanding into RL, aiming to build an RL framework that can acquire knowledge from texts and integrate it into the underlying decision-making problems. [11, 12] by proposing semi-supervised RL methods to address reward sparsity problem and manual feature engineering. Since the beginning, RL algorithms have been applied more frequently to problems with a finite set of actions [13, 14]. However, continuous RL problems [4, 15, 16] have also become an interesting research subject.

In Markovian decision-making problems (MDPs) based on the Bellman principle of optimality, assuming that we have an accurate model of the agent-environment interactions, the problem can be solved recursively. However, due to the fact that most of the time, such a model barely exists. Therefore model-free methods are developed to solve realistic problems. Based on the policy representation, model-free RL methods can be categorized into value-based and policy-based methods [17]. In the policy-based RL methods, an explicit representation of the policy is presented to the agent during the learning process, whereas in the value-based methods, the policy is implicitly driven by updating the value function estimates. Off-policy value-based methods are inherently more sample efficient than policy-based methods [18].

Exploration and exploitation dilemma along with the sample efficiency are two significant challenges in the RL. A good RL algorithm should balance the requirement to learn more about the environment (exploration) against the desire to perform well based on the currently available knowledge of the environment (exploitation) [19]. One of the widely used exploration schemes is $\varepsilon$-greedy algorithm [20]. This algorithm selects the best action with probability $1-\varepsilon$, $(\varepsilon \in (0,1))$, or choses a random action with probability $\varepsilon$. The $\varepsilon$-greedy method is simple and effective, but it has one main drawback, which is that during the exploration, it chooses equally among all actions. A better alternative for the $\varepsilon$-greedy exploration scheme is to use instead of a uniform distribution, the Boltzmann distribution for selecting an action during exploration, which is exactly the case in softmax action selection strategy. In the softmax exploration strategy, a positive parameter $\tau$ called temperature is responsible for balancing the exploration against exploitation. By choosing $\tau = 0$, the agent holds the exploration, while as $\tau \to \infty$, the agent tends to behave randomly during the exploration.

Utilizing human knowledge has been studied before in many scientific domains including RL [21-26]. Inspired by the way human beings learn from the knowledge of an expert, in this paper we proposed a state-of-the-art value-based action selection method which explicitly incorporates human knowledge into RL algorithm via a fuzzy logic system. The overall method consists of action-value estimation, a fuzzy mapping, and an action selection strategy based on the output of the fuzzy mapping. Furthermore, a tunable parameter is defined as a control parameter which can be interpreted as a means to integrate expert's knowledge without the need to define any IF-THEN rule directly by the expert. [27] has worked on a similar concept. In [27], a policy-based method combined with a fuzzy logic system named knowledge controller which takes environments states as inputs is proposed. In contrast to [27], we adopted a different approach for human knowledge integration in RL. We proposed a value-based RL method with the fuzzy logic system as a transformation between estimated action-values and action selection policy. For the proof of concept, the n-armed bandit problem is used as a test-bed to show the method.

The article is organized as follows: in section II, we deal with the n-armed bandit problem formulation and review the general action selection axioms. In section III, we go through the proposed RL algorithm, and finally, in sections IV and V the simulation results and conclusions are presented.

## II. PRELIMINARIES

### A. n-ARM BANDIT PROBLEM
n-Armed bandit problem is an MDP with only one state and finite action space that commonly has been employed for evaluating the exploration and exploitation strategies in RL [1, 28]. Generally, any problem in which we are faced repeatedly with a selection among n different choice of actions associated with numerical rewards based on a probability distribution is an n-armed bandit problem. In its original formulation, we have n slot machines or equivalently one slot machine with n levers, and the rewards are the payoffs for hitting the jackpot.

The objective is to find a sequence of actions that maximize the expected total reward at the end of the course. During the procedure, we can maximize our rewards by focusing more on the best levers until the current step. Considering it as an RL problem, the state-action value is the mean reward of hitting the levers and is assumed to be unknown in this formulation.

Considering the estimation of the action values, then at any step, there is at least one action that maximizes the state-action value until that step, which will be called a greedy action. By exploitation, we mean selecting a greedy action, while exploration is the process of choosing among those non-greedy actions, which enables us to discover whether other actions may lead to better results. Exploration improves our estimation of non-greedy actions and maximizes the long-term rewards, while exploitation ensures the currently expected reward to be maximum.

We use the same sample-average method for estimating state-action values [1]. As defined in previous paragraphs, the true value of an action $a$ is the mean value of the rewards by choosing that specific action. Although it has some exceptions likes in the case of Cauchy distribution [29], the law of large-numbers predicts the value for an action, like $a$, as follow:

$$Q_t(a) = \frac{\sum_{i=1}^{n_a} r_{a_i}}{n_a} \quad (1)$$

in which, we assume that the action $a$ has been selected $n_a$ times among all other possible actions until the current step $t$, and $r_{a_i}$ is the associated reward for choosing the action $a$ for the $i$-th time. We also assume that $Q_{n_a}(a) = 0$, in the cases of $n_a = 0$. In the $t$-th step, the greedy action $a^*$ is $\text{argmax}_a Q_t(a)$.

Choosing the greedy action ensures that in each step, we have the maximum possible reward based on the current knowledge, yet, it does not guarantee that we have the maximum award in the long-term scenarios.

Choosing the greedy action for all the steps except for a small number of them is the $\varepsilon$-greedy approach to the problem. Considering the $\varepsilon$ to be a small positive real number, this is equivalent to picking randomly among the actions, while greedy action has the probability of $1-\varepsilon$, and others having the total probability of $\varepsilon$. As the number of steps increases, this method guarantees that each action has been sampled an infinite number of times, and based on the law of large numbers means, $Q_t(a)$ for each action, $a$, converges to the actual value.

Although $\varepsilon$-greedy strategy is an easy and accessible way of balancing exploration against exploitation in RL, the major drawback is that, during the exploration, it chooses equally among all the actions. Hence, it is as likely to select the worst action as it is to pick the next-to-best.

## B. PROBLEM FORMULATION

An obvious solution for the previous problem of the $\varepsilon$-greedy strategy is to use a better grading function to choose among the actions. The greedy action still gives the highest selection probability, but all the others are ranked and weighted according to their estimated values at the current step. Back to n-armed bandit problem, if we have $n$ machines or equivalently a machine with $n$ levers, with estimated state-action values at time step $t$ recorded as $Q = [Q_t(1), ..., Q_t(n)]^T$, we want to construct the vector $\pi = [\pi_t(1), ..., \pi_t(n)]^T$ which actually is the stochastic policy distribution being in the current state at time step $t$. The probabilities $\pi_t(i), i = 1, ..., n$ satisfy the following axioms:

- For each action $a$, we have $\pi_t(a) \geq 0$.
- For each actions $a$ and $b$ respectively with the current estimates $Q_t(a)$ and $Q_t(b)$ such that $Q_t(a) \geq Q_t(b)$, we have $\pi_t(a) \geq \pi_t(b)$.
- Denoting the set of all the actions as $A$, we have $\sum_{a \in A} \pi_t(a) = 1$.

These axioms imply a transformation between the vector of currently estimated state-action values $Q$ and the vector of current selection probabilities $\pi$. We denote this transformation by $T$. Considering $l$ real numbers $\alpha_1, \alpha_2, ..., \alpha_l$ to be the control parameters, we start by formulating the transformation as a vector of scalar functions for each action $a \in A$ as follow:

$$T(Q_t(a), \alpha_1, ..., \alpha_l) \quad (2)$$

Substituting $T$ with any non-decreasing and positive function would satisfy the first two axioms, while in order to satisfy the third one, the norm 1 of vector $\pi$, which is the sum of the elements of the vector, should be equal to one. We can normalize $\pi$ by dividing each element of it by the sum of all elements. With this modification, each element of the probability vector is:

$$\pi_t(a) = \frac{T(Q_t(a), \alpha_1, \alpha_2, ..., \alpha_l)}{\sum_{a \in A} T(Q_t(a), \alpha_1, \alpha_2, ..., \alpha_l)} \quad (3)$$

As an example, we can use the exponential function in the place of $T$, which by letting $l = 1$ and $\alpha_1 = \tau$ leads to Boltzmann grading among the actions:

$$\pi_t(a) = \frac{e^{\frac{Q_t(a)}{\tau}}}{\sum_{a \in A} e^{\frac{Q_t(a)}{\tau}}} \quad (4)$$

where $\alpha_1 = \tau$ is a positive real constant, called temperature.

As soon as we choose the control parameters such that the first two conditions hold and the functions are well-defined, no other constraints will remain for constructing the transformation function. In order to uniquely define the transformation, the aforementioned axioms are not sufficient. Therefore, there are many such transformations satisfying the three axioms above. For example, considering the exponential functions to form the transformation as in (4), as soon as we choose $\tau$ to be positive, the transformation is well-defined and satisfies all the axioms which means, even in this case, with only one control parameter, the conditions are not enough to uniquely determine the transformation. Hence, having a non-empty feasibility set, it gives us the possibility to solve further optimization problems such as finding the parameters for which we have the fastest possible learning rate or ensuring other required constraints.

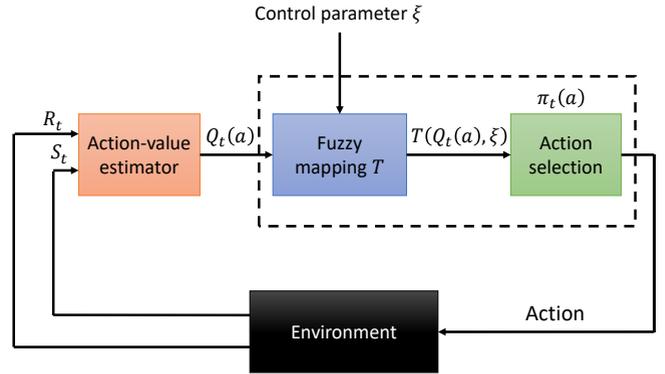

**FIGURE 1.** The block diagram of the proposed action selection method.

## III. PROPOSED METHOD

The capability of fuzzy logic to describe system behaviors using simple conditional IF-THEN rules gives us this ability to employ expert's knowledge in system modeling and control. This potential has created a practical solution for a wide variety of applications in industry, engineering, science, and even human sciences [30-34]. In this section, we use fuzzy logic potential to design a new policy to select actions in the RL problems.

Fuzzy logic as a universal nonlinear function approximator has been used for defining the previously discussed nonlinear transformation [35]. This way, we can use linguistic variables and rules to define the transformation; hence, by implementing a fuzzy logic system in our action selection method, we propose a novel framework for integrating experts' knowledge to the solution.

A fuzzy rule base consists of a set of fuzzy IF-THEN rules and it is the heart of the fuzzy system. Each fuzzy rule base contains the following fuzzy IF-THEN rules:

$$Rule_j: if\ x_1\ is\ A_1^j\ \&\ x_2\ is\ A_2^j\ \&\ ...\ \&\ x_p\ is\ A_p^j\ then\ y\ is\ B^j \quad (5)$$

Where $A_i^j$ and $B^j$ are fuzzy sets in input sets $U_i \subset \mathbb{R}$ and output set $V \subset \mathbb{R}$ respectively, $x = [x_1, x_2, ..., x_p]^T \in U$ and $y \in V$ are the inputs and the output (linguistic) variables of the fuzzy system. $j = 1, 2, ..., m$ denotes the index of the rules where $m$ is the total number of rules in the rule base. Each fuzzy set $A_i^j$ is associated with a membership function $\mu_{A_i^j}(y)$. Output sets are also associated with the membership functions $\mu_{B^j}(y)$. Using product inference, singleton fuzzifier, and center average defuzzifier, the fuzzy system describes the following nonlinear mapping [36]:

$$T(x) = \frac{\sum_j^m o^j \left( \prod_i^p \mu_{A_i^j}(x_i) \right)}{\sum_j^m \left( \prod_i^p \mu_{A_i^j}(x_i) \right)} \quad (6)$$

In this note, we consider a single-input single-output fuzzy system ($p = 1$) and set $m = n$, where $n$ is the number of actions in the RL problem. Using Gaussian membership functions for both input and output, we have:

$$\mu_{A^j}(x) = e^{-\left(\frac{x - I^j}{\sigma_x^j}\right)^2}, \quad \mu_{B^j}(y) = e^{-\left(\frac{y - O^j}{\sigma_y^j}\right)^2} \quad (7)$$

Based on which we can define the following nonlinear mapping:

$$T(Q_t(a), \xi) = \frac{\sum_{j=1}^n O(\xi, j) e^{-\left(\frac{Q_t(a) - I^j}{\sigma_x}\right)^2}}{\sum_{j=1}^n e^{-\left(\frac{Q_t(a) - I^j}{\sigma_x}\right)^2}} \quad (8)$$

Assuming $Q_t(.)$ takes value in the interval $[\alpha, \beta]$ at each play (step) of the learning process, we define $I^j$ such that the input membership functions uniformly cover this interval. Therefore, we have:

$$I^j = \frac{\alpha(n-j) + \beta(j-1)}{n-1} \quad (9)$$

We want the fuzzy sets to be roughly consistent, and hence, the input variance is empirically selected to be:

$$\sigma_x = \frac{\beta - \alpha}{3n - 1} \quad (10)$$

The output variance $\sigma_y$ has also been chosen empirically as follows:

$$\sigma_y = \frac{1}{2n - 1} \quad (11)$$

The function $O(\xi, j)$ determines the center of output membership functions:

$$O(\xi, j): \begin{cases} \max\left\{1 - \xi, \frac{1}{2} e^{-\left(\frac{\xi - 0.5}{0.15\sqrt{2}}\right)^2}\right\}, & j = n \\ \frac{j-1}{2(n-1)} e^{-\left(\frac{\xi - 0.5}{0.15\sqrt{2}}\right)^2}, & j < n \end{cases} \quad (12)$$

The main tuning parameter in this system is $\xi$, which belongs to the interval $[0,1]$, by which we can change the agent's view of the action selection policy. Changing $\xi$ only affects the separation of the output membership functions. High $\xi$ (near to 1) causes the actions to be almost equiprobable, thus agent performs exploration. However, low $\xi$ (near to 0) leads to a more significant difference in selection probability for actions that differ in their estimated values and consequently, the agent performs exploitation.

By tuning a proper value for control parameter $\xi$ and using the function $O(\xi, j)$, we can define a fuzzy mapping with modifiable rule base that takes the estimated action-values as inputs. This mapping satisfies the previously discussed axioms of being a positive and non-decreasing function. In Fig. 1 the block diagram of the proposed framework is shown. This diagram is composed of an action-value estimator, a fuzzy mapping, and an action selection block. We choose sample-average method for action-value estimation. In the action selection block, we can use the same procedure for the normalization as discussed in part B of section II. Finally, the fuzzy mapping and the action selection blocks together satisfy all three axioms.

## IV. SIMULATION RESULTS

Assuming $n = 10$ the output membership functions have been shown in Fig. 2, Fig. 3, Fig. 4, and Fig. 5, which are associated with $\xi = 0$, $\xi = 1$, $\xi = 0.6$, $\xi = 0.75$.

The proposed method with $\xi = 0.04$ has been tested over 1000 runs of 10-arm bandit problem, and the results were compared to softmax method applied to the same problem with the temperature chosen to be $\tau = 0.1$. The tuning parameters for both methods were changed over the possible ranges, and the chosen parameters are associated with both methods' optimal behaviors.

As shown in Fig. 6, the softmax will reach the optimal behavior after passing approximately 50% of the tasks, while the proposed method settled after passing approximately only 20% of the tasks. The maximum, mean, and median of each method's results have been compared in Table I to show the significance of the difference between these two methods. As shown in Table I, all of these three parameters in the proposed method are larger than the softmax method. Besides, the difference between maximum and median in the proposed method is less than softmax, which for the n-armed bandit problem, means that we can obtain more rewards by fewer plays compared to using the softmax strategy.

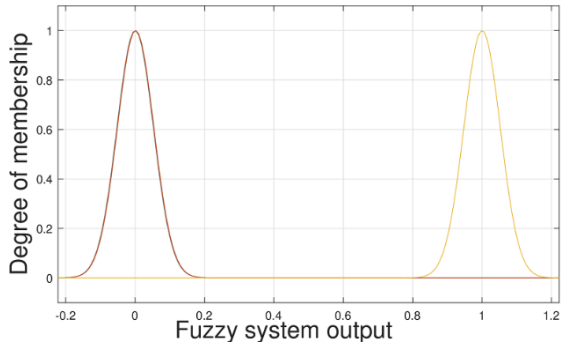

**FIGURE 2.** Output membership functions for $\xi = 0$.

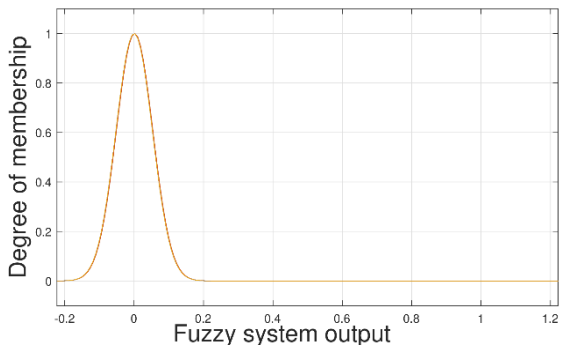

**FIGURE 3.** Output membership functions for $\xi = 1$.

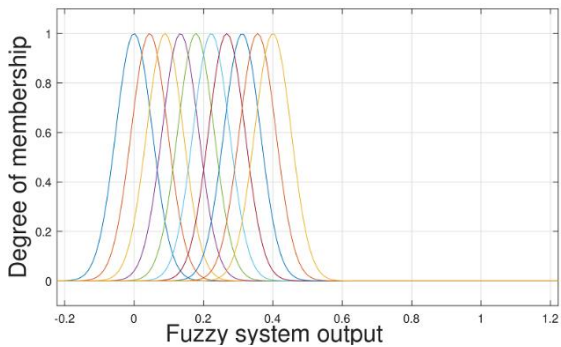

**FIGURE 4.** Output membership functions for $\xi = 0.6$.

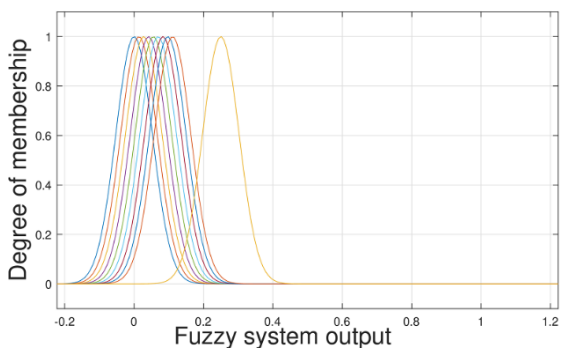

**FIGURE 5.** Output membership functions for $\xi = 0.75$.

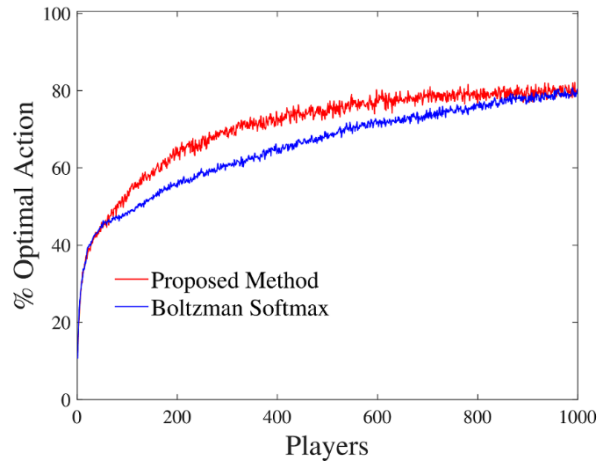

**FIGURE 6.** Percentage of optimal action in n-armed bandit with 1000 runs for both the proposed method and softmax method.

TABLE I.
COMPARISON OF STATISTICS

| Statistical parameters | SOFTMAX | Proposed method |
|---|---|---|
| **Maximum** | 80.7 | 82.1 |
| **Mean** | 68.3 | 70.5 |
| **Median** | 65.5 | 75.0 |
| **Maximum-Median** | 15.1 | 7.1 |

## V. Conclusion

A new action selection policy defined by a nonlinear fuzzy mapping has been proposed for reinforcement learning problems. The simulation results confirm that the proposed method has a better performance and convergence rate than the existing alternatives such as the Boltzmann softmax method, while at the same time using fuzzy logic, we can better integrate the expert's knowledge into the downstream decision-making problem; Thus, this method is more intuitive and more interpretable than other similar action selection strategies. Future works can be inspired by the mapping concept with control parameters proposed here. For example, adding more control parameters or using different functions to relate control parameters to membership functions might be good options to improve the results. In addition, testing the proposed method on a variety of test-beds and developing the proposed concept on continuous action space could also be considered as future works.